\documentclass[11pt]{article}
\usepackage{coling2018}
\usepackage{times}
\usepackage{url}
\usepackage{latexsym}
\usepackage{amsmath}
\usepackage{multirow}
\usepackage{mathrsfs}
\usepackage{enumitem}
\usepackage{graphics,graphicx}
\usepackage{algorithm}
\usepackage{caption}
\usepackage{amssymb}
\usepackage[noend]{algpseudocode}
\usepackage{esvect}
\usepackage{booktabs}
\usepackage{placeins}
\usepackage[toc,page]{appendix}
\usepackage{diagbox}
\usepackage{pdflscape}
\usepackage{longtable}
\usepackage{subcaption}

\title{Can Taxonomy Help? Improving Semantic Question Matching using Question Taxonomy}
\author{Deepak Gupta$^{\ast}$, Rajkumar Pujari$^{\dagger}$, Asif Ekbal$^{\ast}$, Pushpak Bhattacharyya$^{\ast}$,\\  \textbf{Anutosh Maitra$^{\ddagger}$} \textbf{Tom Jain$^{\ddagger}$} and \textbf{Shubhashis Sengupta$^{\ddagger}$} \\
  $^{\ast}$Indian Institute of Technology Patna, India  \\
 $^{\dagger}$Purdue University, USA \\
  $^{\ddagger}$Accenture Labs, Bengaluru, India \\
  {\tt $^{\ast}$\{deepak.pcs16,asif,pb\}@{iitp.ac.in}, $^{\dagger}$rpujari@purdue.edu }
  \\{\tt $^{\ddagger}$\{anutosh.maitra,tom.geo.jain,shubhashis.sengupta\}@{accenture.com}}}

\date{}

\begin{document}
\maketitle
\begin{abstract}
In this paper, we propose a hybrid technique for semantic question matching. It uses our proposed two-layered taxonomy for English questions by augmenting state-of-the-art deep learning models with question classes obtained from a deep learning based question classifier. Experiments performed on three open-domain datasets demonstrate the effectiveness of our proposed approach. We achieve state-of-the-art results on partial ordering question ranking (POQR) benchmark dataset. Our empirical analysis shows that coupling standard distributional features (provided by the question encoder) with knowledge from taxonomy is more effective than either deep learning (DL) or taxonomy-based knowledge alone.
\end{abstract}
\blfootnote{
    
    %
    This work is licensed under a Creative Commons 
    Attribution 4.0 International Licence.
    Licence details:
    \url{http://creativecommons.org/licenses/by/4.0/}.
    %
    %
}
\section{Introduction} \label{sec:intro}
\noindent Question Answering (QA) is a well investigated research area in Natural Language Processing (NLP). There are several existing QA systems that answer factual questions with short answers \cite{iyyer2014neural,bian2008finding,ng2015qanus}. However, systems which attempt to answer questions that have long answers with several well-formed sentences, are rare in practice. This is mainly due to some of the following challenges: (i) selecting appropriate text fragments from document(s), (ii) generating answer texts with coherent and cohesive sentences, (iii) ensuring the syntactic as well as semantic well-formedness of the answer text. However, when we already have a set of answered questions, reconstructing the answers for semantically similar questions can be bypassed. For each unseen question, the most semantically similar question is identified by comparing the unseen question with the existing set of questions. The question, which is closest to the unseen question can be retrieved as a possible semantically similar question. Thus, accurate semantic question matching can significantly improve a QA system. In the recent past, several deep learning based models such as recurrent neural networks (RNNs), convolution neural network (CNN), gated recurrent units (GRUs) etc. have been explored to obtain representation at the word \cite{mikolov2013efficient,pennington2014glove}, sentence \cite{kimCNN} and paragraph \cite{zhang2017deconvolutional} level.\\
\indent In the proposed semantic question matching framework, we use attention based neural network models to generate question vectors. We create a hierarchical taxonomy by considering different types and subtypes in such a way that questions having similar answers belong to the same taxonomy class. We propose and train a deep learning based question classifier network to classify the taxonomy classes. The taxonomy information is helpful in taking a decision on semantic similarity between them. For example, the questions `\textit{How do scientists work?}' and `\textit{Where do scientists work?}', have very high lexical similarity but they have different answer types. This can be easily identified using a question taxonomy. Taxonomy can provide very useful information when we do not have enough data for generating useful deep learning based representations, which are generally the case with restricted domains. In such scenarios linguistic information obtained from the prior knowledge helps significantly in improving the performance of the system.\\
\indent We propose a neural network based algorithm to classify the questions into appropriate taxonomy class(es). The information, thus obtained from taxonomy, is used along with the DL techniques to perform semantic question matching. Empirical evidence establishes that our taxonomy, when used in conjunction with Deep Learning (DL) representations, improves the performance of the system on semantic question (SQ) matching task.\\
\indent We summarize the contributions of our work as follows: \textbf{(i)} We create a two-layered taxonomy for English questions; \textbf{(ii)} We propose a deep learning based method to identify taxonomy classes of questions; \textbf{(iii)} We propose a dependency parser based technique to identify the \textit{focus} of the question; \textbf{(iv)} We propose a framework to integrate semantically rich taxonomy classes with DL based encoder to improve the performance and achieve new state-of-the-art results in semantic question ranking on benchmark dataset and Quora dataset; and finally \textbf{(v)} We release two annotated datasets, one for semantically similar questions and the other for question classification.
\section{Related Works} \label{section:related}
\noindent Rapid growth of community question and answer (cQA) forums have intensified the necessity for semantic question matching in QA setup. Answer retrieval of semantically similar questions has drawn the attention of researchers in very recent times \cite{marquez2015semeval,nakov16lluis}. It solves the problem of \textit{question starvation} in cQA forums by providing a semantically similar question which has already been answered. In literature, there have been attempts to address the problem of finding the most similar match to a given question, for e.g. \newcite{burke1997question} and \newcite{mlynarczyk2005faqfinder}. \newcite{wang2009syntactic} have presented syntactic tree based matching for finding semantically similar questions. `Similar question retrieval' has been modeled using various techniques such as topic modeling \cite{li2011improving}, knowledge graph representation \cite{zhou2013improving} and machine translation \cite{jeon2005finding}. Semantic kernel based similarity methods for QA have also been proposed in \cite{Filice2016KeLPAS,Croce2017DeepLI,Croce2011StructuredLS}.\\
\indent Answer selection in QA forums is similar to the question similarity task. In recent times, researchers have been investigating DL-based models for answer selection \cite{wang-nyberg:2015:ACL-IJCNLP,severyn2015learning,feng2015applying}. Most of the existing works either focus on better representations for questions or linguistic information associated with the questions. On the other hand, the model proposed in this paper is a hybrid model. We also present a thorough empirical study of how sophisticated DL models can be used along with a question taxonomy concepts for semantic question matching.
\section{Question Matching Framework}\label{sec:frame}
When framed as a computational problem, semantic question (SQ) matching for QA becomes equivalent to ranking questions in the existing question-base according to their semantic similarity to the given input question. Existing state-of-the-art systems use either deep learning models \cite{lei-EtAl:2016:N16-1} or traditional text similarity methods \cite{jeon2005finding,wang2009syntactic} to obtain the similarity scores. In contrast, our framework of SQ matching efficiently combines deep learning based question encoding and a linguistically motivated taxonomy. Algorithm \ref{algo:qq_match} describes the precise method we follow. $Similarity(.)$ is the standard cosine similarity function. $fsim$ is focus embedding similarity which is described later in Section \ref{experiment:setup}.
\begin{figure}
\centering
\begin{minipage}{.8\linewidth}
\begin{algorithm}[H]
\begin{algorithmic}
\Procedure{SQ Matching}{QSet}
\State RESULTS $\gets$ \{\}
\For{($p$, $q$) in QSet}
\State $\vec{p}, \vec{q}$ $\gets$ \textit{Question-Encoder}($p$, $q$)
\State $sim$ $\gets$ \texttt{Similarity}($\vec{p}$, $\vec{q}$)
\State ${T_p^{c}, T_q^{c}}$ $\gets$ \texttt{Taxonomy-Classes}($p$, $q$)
\State $F_p, F_q$ $\gets$ \texttt{Focus}($p$, $q$)
\State $\vec{F_p}, \vec{F_q}$ $\gets$ \texttt{Focus-Encoder}($F_p,F_q$)
\State $fsim$ $\gets$ \texttt{Similarity}($\vec{F_p}$, $\vec{F_q}$)
\State Feature-Vector=[$sim$, ${T_p}^{c}$, ${T_q}^{c}$, $fsim$]
\State {result $\gets$ Classifier(Feature-Vector)}
\State RESULTS.append(result)
\EndFor
\Return RESULTS
\EndProcedure
\end{algorithmic}
\caption{Semantic Question Matching}
\label{algo:qq_match}
\end{algorithm}
\end{minipage}
\end{figure}
\subsection{Question Encoder Model} \label{subsec:QEModel}
Our question encoder model is inspired from the state-of-the-art question encoder architecture proposed by \newcite{lei-EtAl:2016:N16-1}. We extend the question encoder model of \newcite{lei-EtAl:2016:N16-1} by introducing attention mechanism similar to \newcite{bahdanau2014neural} and \newcite{chopra2016abstractive}. We propose the attention based version of two question encoder models, namely Recurrent Convolutional Neural Network (RCNN) \cite{lei-EtAl:2016:N16-1} and Gated Recurrent Unit (GRU)  \cite{chung2014empirical,cho2014properties}. \\
\indent A question encoder with attention does not need to capture the whole semantics of the question in its final representation. Instead, it is sufficient to capture a part of hidden state vectors of another question it needs to attend while generating the final representation.
Let \textbf{H}$ \in \mathbb{R}^{d \times n}$ be a matrix consisting of hidden state vectors $[h_1, h_2 \ldots h_n]$ that the question encoder (RCNN, GRU) produced when reading the $n$ words of the question, where $d$ is a hyper parameter denoting the size of embeddings and hidden layers. The attention mechanism will produce an attention weight vector $\alpha_t \in \mathbb{R}^n$ and a weighted hidden representation $r_t \in \mathbb{R}^d$. 
\begin{equation}
\begin{aligned}
C_t&=tanh(W^{H}H+W^{v}(v_t \otimes I_n))\\
\alpha_t & = softmax(w^TC_t)\\
r_t&=H\alpha^T
\end{aligned}
\end{equation}
where $W^H$, $W^v \in \mathbb{R}^{d\times d}$,
are trained projection matrices. $w^T$ is the transpose of the trained vector $w \in \mathbb{R}^d$. $v_t \in \mathbb{R}^{d}$ shows the embedding of token $x_t$ and $I_n \in \mathbb{R}^{n} $ is the vector of $1$. The product $W^{v}(v_t \otimes I_n)$ is repeating the linearly transformed $v_t$ as many times (n) as there are words in the candidate question. Similarly we can obtain the attentive hidden state vectors $[r_1, r_2 \ldots r_n]$. We apply the averaging pooling strategy to determine the final representation of the question. 

\indent Annotated data, $\mathscr{D}=\{(q_i,p_i^+,p_i^-)\}$ is used to optimize $f(p,q,\phi)$, where $f(.)$ is a measure of similarity between the questions $p$ and $q$, and $\phi$ is a parameter to be optimized. Here $p_i^+$ and $p_i^-$ correspond to the similar and non-similar question sets, respectively for question $q_i$. Maximum margin approach is used to optimize the parameter $\phi$. For a particular training example, where $q_i$ is similar to $p_i^+$, we minimize the max-margin loss $\mathcal{L}(\phi)$ defined as:
\begin{equation}
\footnotesize
\mathcal{L}(\phi)=\max_{ p \in Q^{'}(q_i) }\big\{f(q_i,p;\phi)-f(q_i, p_i^+;\phi)+\lambda(p,p_i^+) \big\}
\end{equation}
where $Q^{'}(q_i)=p_i^{+} \cup p_i^{-}$, $\lambda(p,p_i^+)$ is a positive constant set to $1$ when $p \neq p_i^+$, $0$ otherwise.
\subsection{Question Taxonomy} \label{sec:taxonomy}
\noindent Questions are ubiquitous in natural language. Questions essentially differ on two fronts: semantic and syntactic. Questions that differ syntactically might still be semantically equivalent. Let us consider the following two questions:
\begin{itemize}[nosep]
    \item{What is the number of new hires in 2018?}
    \item{How many employees were recruited in 2018?}
\end{itemize}
\indent Although the above questions are not syntactically similar but both are semantically equivalent and have the same answer. A well-formed taxonomy and question classification scheme can provide this information which eventually helps in determining the semantic similarity between the questions.\\
\indent According to \newcite{gruber1995}, ontologies are commonly defined as specifications of shared conceptualizations. Informally, conceptualization is the relevant informal knowledge one can extract from their experience, observation or introspection. Specification corresponds to the encoding of this knowledge in representation language. In order to create a taxonomy for questions, we observe and analyze questions from Stanford Question Answering Dataset (SQuAD) released by \newcite{squad2016} and question classifier data from \newcite{hovy2001toward} and \newcite{liroth2002}. The SQuAD dataset consists of 100,000+ questions and their answers, along with the text extracts from which the questions were formed. The other question classifier dataset contains $5,500$ questions. In the succeeding subsections, we describe in detail, the coarse classes, fine classes and focus of a question. We have included an additional hierarchical taxonomy table with one example question for each class in the appendix.
\subsubsection{Coarse Classes} To choose the correct answer of a question one needs to understand the question and categorize the answer into the appropriate category which could vary from a basic implicit answer (question itself contains the answer) to a more elaborate answer (description). The coarse class of question provides a broader view of the expected answer type. We define the following six coarse class categories: \textit{Quantification, Entity, Definition, Description,} \textit{List} and \textit{ Selection}. \textit{Quantification} class deals with the questions which look for a specific quantity as answer. Similarly \textit{Entity, Definition, Description} classes give evidence that the answer type will be entity, definition or a detailed description, respectively. 
\textit{Selection} class defines the question that looks for an answer which needs to be selected from the given set of answers. Few examples of questions along with their coarse class are listed here:
\begin{itemize}[noitemsep, topsep=0pt]
 \item \textbf{Quantity:} \textit{Give the average speed of 1987 solar powered car winner?}
 \item \textbf{Entity:}\textit{ Which animal serves as a symbol throughout the book?}
\end{itemize}

\subsubsection{Fine Classes}
Coarse class defines the answer type at the broad level such as entity, quantity, description etc. But extracting the actual answer of question needs further classification into more specific answer types. Let us consider the following examples of two questions:
\begin{enumerate}[noitemsep, topsep=0pt]
\item \textbf{Entity (Flora):} \textit{What is one aquatic plant that remains submerged?}
\item \textbf{Entity (Animal):} \textit{Which animal serves as a symbol throughout the book?}
\end{enumerate}
Although both the questions belong to the same coarse class \textit{entity} but they belong to the different fine classes, (\textit{Flora} and \textit{Animal}). Fine class of a question is based on the nature of the expected answer. It is useful in restricting the potential candidate matches. Although, questions belonging to the same fine class need not to be semantically same, questions belonging to the different fine classes rarely match. 
We show the set of the proposed coarse class and their respective fine classes in Table \ref{table:tax}.
\begin{table}[]
\centering
\resizebox{0.7\linewidth}{!}{%
\begin{tabular}{l|l}
\hline
\textbf{Coarse Classes} & \multicolumn{1}{c}{\textbf{Fine Classes}} \\ \hline
\textbf{Quantification} & \begin{tabular}[c]{@{}l@{}}Temprature, Time/Duration, Mass, Number, Age\\ Distance, Money, Speed, Size, Percent, Rank/Rating\end{tabular} \\ \hline
\textbf{Entity} & \begin{tabular}[c]{@{}l@{}}Person, Location, Organization, Animal, Technique\\ Flora, Entertainment, Food, Abbreviation, Language\\ Disease, Award/Title, Event, Sport/Game, Policy, Date\\ Publication, Body, Thing, Feature/Attribute, Website\\ Industry Sector, Monuments, Activity/Process, Other\\ Tangible, Other Intangible\end{tabular} \\ \hline
\textbf{Definition} & Person, Entity \\ \hline
\textbf{Description} & \begin{tabular}[c]{@{}l@{}}Reason, Mechanism, Cause \& Effect, Describe\\ Compare \& Contrast, Analysis\end{tabular} \\ \hline
\textbf{List} & \begin{tabular}[c]{@{}l@{}}Set of fine classes listed in the coarse classes \\ \textit{Quantification} and \textit{Entity}\end{tabular} \\ \hline
\textbf{Selection} & Alternative/Choice, True/False \\ \hline
\end{tabular}%
}
\caption{Set of proposed coarse and respective fine classes}
\label{table:tax}
\end{table}
\subsubsection{Focus of a Question}
According to \newcite{moldovan2000}, \textit{focus} of a question is a word or a sequence of words, which defines the question and disambiguates it to find the correct answer the question is expecting to retrieve. In the following example, \textit{Describe the customer service model for Talent and HR BPO}, the term `model' serves as the \textit{focus}. As per \newcite{bunescu2010}, \textit{focus} of a question is contained within the noun phrases of a question. In the case of imperatives, the direct object (\textit{dobj}) of the \textit{question word} contains the \textit{focus}. Similarly, in case of interrogatives, there are certain dependencies that capture the relation between the question word and its focus. The \textit{dobj} relation of the root verb or \textit{det} relation of \textit{question word} for interrogatives contain the \textit{focus}. Question word \textit{how} has \textit{advmod} relations that contain \textit{focus} of the question. Priority order of the relations used to extract \textit{focus} is obtained by observation on the SQuAD data. We depict the pseudo-code of the \textit{focus} extraction method in the appendix.
\subsection{Question Classification}
Question classification guides a QA system to extract appropriate candidate answer from the document/corpus. For example, the question \textit{`How much does international cricket player get paid?'} should be accurately classified as the coarse class \textit{quantification} and fine class \textit{money} to further extract the appropriate answer. In our problem, we attempt to exploit the taxonomy information to identify the semantically similar questions. Therefore, the question classifier should be capable enough to accurately classify the coarse and fine classes of a reformulated question:
\begin{enumerate}[noitemsep, topsep=0pt]
\item \textit{What is the salary of an international level cricketer?}
\item \textit{What is the estimated wage of an international cricketer?}
\end{enumerate}
\subsubsection{Question Classification Network}
In order to identify the coarse and fine classes of a given question, we employ a deep learning based question classifier. In our question classification network  CNN and bidirectional GRU has been applied sequentially. The obtained question vector is passed through a feed forward NN layer, and then through a softmax layer to obtain the final class of the question. We use two separate classifiers for coarse and fine class classification.\\
\indent Firstly, an embedding layer maps a question $Q=[w_1, w_2 \ldots w_n]$, which is a sequence of words $w_{i}$, into a sequence of dense, real-valued vectors,  $E=[v_1, v_2 \ldots v_n]$, $v_{i} \in \mathbb{R}^d$. Thereafter, a convolution operation is performed over the zero-padded sequence $E^{p}$. $F \in \mathbb{R}^{k \times m \times d}$, a set of $k$ filters is applied to the sequence. We obtain convoluted features $c_t$ at given time $t$ for $t = 1, 2, \ldots{}, n$.
\begin{equation}
c_t=tanh(F[v_{t-\frac{m-1}{2}} \ldots v_t \ldots v_{t+\frac{m-1}{2}}])
\end{equation}
Then, we generate the feature vectors $C^{\prime}=[c_1^{\prime}, c_2^{\prime} \ldots c_n^{\prime}]$, by applying max pooling on $C$. This sequence of convolution feature vector $C^{\prime}$ is passed through a bidirectional GRU network. We obtain the forward hidden states $\overrightarrow{h_t}$ and backward hidden states $\overleftarrow{h_t}$ at every step time $t$. The final output of recurrent layer $h$ is obtained as the concatenation of the last hidden states of forward and backward hidden states.\\
\indent Finally, the fixed-dimension vector $h$ is fed into the softmax classification layer to compute the predictive probability $p(y=l|Q)=\frac{exp(w_{l}^Th+b_l)}{\sum_{i=1}^{L}exp(w_{i}^Th+b_i)}$ for all the question classes (coarse or fine). We assume there are $L$ classes where $w_{x}$ and $b_{x}$ denote the weight and bias vectors, respectively and $x \in \{l,i\}$. 
\subsection{Comparison with Existing Taxonomy}
In the Text REtrieval Conference (TREC) task, \newcite{liroth2002} proposed a taxonomy to represent a natural semantic classification for a specific set of answers. This was built by analyzing the TREC questions. In contrast to \newcite{liroth2002}, along with TREC questions we also make a thorough analysis of the most recent question answering dataset (SQuAD) which has a collection of more diversified questions. Unlike \newcite{liroth2002}, we introduce the list and selection type question classes in our taxonomy. Each of these question types has its own strategy to retrieve an answer, and therefore, we put these separately in our proposed taxonomy. The usefulness of list as a different coarse class in semantic question matching can be understood considering the following questions:
\begin{enumerate}[noitemsep, topsep=0pt]
\item \textit{What are some techniques used to improve crop production?}
\item \textit{What is the best technique used to improve crop production ?}
\end{enumerate}
These two questions are not semantically similar as \textbf{(1)} and \textbf{(2)} belong to \textit{list} and \textit{entity} coarse classes, respectively. Moreover, \newcite{liroth2002}'s taxonomy has overlapping classes (\textit{Entity, Human and Location}). In our taxonomy we put all these classes in a single coarse class named \textit{Entity}, which helps in identifying semantically similar questions better. We propose a set of coarse and respective fine classes with more coverage compared to \newcite{liroth2002}. \newcite{liroth2002} taxonomy does not cover many important fine classes such as, \textit{entertainment, award/title, activity, body} etc., under \textit{entity} coarse class. We include these fine classes in our proposed taxonomy. We further redefine \textit{description} type questions by introducing \textit{cause \& effect, compare and contrast} and \textit{analysis} fine classes in addition to \textit{reason, mechanism} and \textit{description} classes. This finer categorization helps in choosing a more appropriate answer strategy for descriptive questions.
\section{Experiments} \label{sec:exp}
\subsection{Datasets} \label{sec:dataset}
\indent We perform experiments on three benchmark datasets,  namely Partial Ordered Question Ranking (POQR)-Simple, POQR-Complex \cite{bunescu2010learning} and Quora datasets. In addition to this, we also perform experiments on a new semantic question matching dataset (Semantic SQuAD\footnote{All the datasets used in the paper are publicly available at \url{https://figshare.com/articles/Semantic_Question_Classification_Datasets/6470726}}) created by us. In order to evaluate the system performance, we perform experiments in two different settings. The first setting deals with semantic question ranking (SQR) and the second deals with semantic question classification (SQC) with two classes (match and no-match). We perform SQR experiments on Semantic SQuAD and POQR datasets. For SQC experiments, we use Semantic SQuAD and Quora datasets. 
\subsubsection{Semantic SQuAD} \label{exp:data:squad}
We built a semantically similar question-pair dataset based on a portion of SQuAD data. SQuAD, a crowd-sourced dataset, consists of 100,000+ answered questions along with the text from which those question-answer pairs were constructed. We randomly selected $6,000$ question-answer pairs from SQuAD dataset and for a given question we asked $12$ annotators\footnote{The annotators are the post-graduate students having proficiency in English language.} to formulate semantically similar questions referring to the same answers. Each annotator was asked to formulate $500$ questions. We divided this dataset into training, validation and test sets of $2,000$ pairs each. We further constructed $4,000$ \textit{semantically dissimilar} questions automatically. We use these $8,000$ question pairs ($4,000$ semantic similar questions pair from test and validation + $4,000$ semantically dissimilar pairs) to train the semantic question classifier for the SQC setting of the experiments. \textit{Semantically dissimilar} questions are created by maintaining the constraint that questions should be from the different taxonomy classes. We perform 3-fold cross-validation on these $8,000$ question pairs. 
\subsubsection{POQR Dataset} \label{exp:data:poqr}
POQR dataset consists of $60$ groups of questions, each having a reference question that is associated with a partially ordered set of questions. Each group has three different sets of questions named as \textit{paraphrase} ($\mathcal{P}$), \textit{useful} ($\mathcal{U}$) and \textit{neutral} ($\mathcal{N}$). For each given reference question $q_r$ we have $q_p \in \mathcal{P}$, $q_u \in \mathcal{U}$, and $q_n \in \mathcal{N}$. As per \newcite{bunescu2010learning} the following two relations hold: 
\begin{enumerate}[noitemsep, topsep=0pt]
\item $(q_p \succ q_u|q_r)$: A \textit{paraphrase} question is \textit{`more useful than'} useful question.
\item $(q_u \succ q_n|q_r)$: A \textit{useful} question is \textit{`more useful than'} neutral question.
\end{enumerate}
By transitivity, it was assumed by \newcite{bunescu2010learning} that the following ternary relation holds $(q_p \succ q_n|q_r)$: ``A \textit{paraphrase} question is \textit{`more useful than'} a neutral question''. We show the statistics of these datasets for \textit{Simple} and \textit{Complex} question types for two annotators (1, 2) in Table \ref{dataset:poqr}. 
\begin{table}[]
\centering
\resizebox{.7\linewidth}{!}{%
\begin{tabular}{c|c|c|c|c}
\hline
\multirow{2}{*}{Datasets} & \multicolumn{2}{c|}{Simple} & \multicolumn{2}{c}{Complex} \\ \cline{2-5} 
 & Simple-1 & Simple-2 & Complex-1 & Complex-2 \\ \hline
$\mathcal{P}$ & 164 & 134 & 103 & 89 \\ \hline
$\mathcal{U}$ & 775 & 778 & 766 & 730 \\ \hline
$\mathcal{N}$ & 594 & 621 & 664 & 714 \\ \hline
Pairs & 11015 & 10436 & 10654 & 9979 \\ \hline
\end{tabular}
}
\caption{Brief statistics of POQR datasets}
\label{dataset:poqr}
\end{table}
\subsubsection{Quora Dataset}\label{exp:dataset:quora}
We perform experiments on semantic question matching dataset consisting of 404,290 pairs released by Quora\footnote{https://data.quora.com/First-Quora-Dataset-Release-Question-Pairs}. The dataset consists of 149,263 matching pairs and 255,027 non-matching pairs. 
\subsection{Evaluation Scheme} \label{exp:eval}
We employ different evaluation schemes for our SQR and SQC evaluation settings. For the \textbf{Semantic SQuAD} dataset, we use the following metrics for ranking evaluation: Recall in top-k results (Recall@$k$) for $k = 1$, $3$ and $5$, Mean Reciprocal Rank (MRR) and Mean Average Precision (MAP). The set of all candidate questions in $2,000$ pairs of the test set is ranked against each input question. As we have only $1$ correct match out of $2,000$ questions for each question in the test set, recall@1 is equivalent to precision@1. Given that we only have one relevant result for each input question, MAP is equivalent to MRR. We evaluate the semantic question classification performance in terms of accuracy. To ensure fair evaluation, we keep the ratio of semantically similar and dissimilar questions to be 1:1. In order to compare the performance on \textbf{POQR dataset} with the state-of-the art results, we followed the same evaluation scheme as described in \newcite{bunescu2010learning}. It is measured in terms of 10-fold cross validation accuracy on the set of ordered pairs, and the performance is averaged between the two annotators (1,2) for the Simple and Complex datasets. For \textbf{Quora dataset}, we perform 3-fold cross validation on the entire dataset evaluating based on the classification accuracy only. We did not perform the semantic question ranking (SQR) experiment on Quora dataset as 149,263 $\times$ 149,263 ranking experiment for matching pairs takes a very long time.
\subsection{Baselines} \label{baseline}
We compare our proposed approach to the following information retrieval (IR) based baselines:\\
\noindent \textbf{1) TF-IDF:} The candidate questions are ranked using cosine similarity value obtained from the TF-IDF  based vector representation.\\
\noindent \textbf{2) Jaccard Similarity:} The questions are ranked using Jaccard similarity calculated for each candidate question with the input question.\\ 
\noindent \textbf{3) BM-25:} The candidate questions are ranked using BM-25 score, provided by Apache Lucene \footnote{\url{https://lucene.apache.org/core/}}.

\subsection{Experimental Setup} \label{experiment:setup}
\textbf{Question Encoder:} We train two different question encoders (hidden size=300) on \textit{Semantic SQuAD} and \textit{Quora} datasets. For Semantic SQuAD dataset, we used $2,000$ training pairs to train the question encoder, as mentioned in Section \ref{exp:data:squad}. For Quora dataset we randomly selected $74,232$ semantically similar question pairs to train the encoder, and $10,000$ question pairs for validation. The best hyper-parameters for the deep learning based attention encoder are identified on validation data. Adam \cite{kingma2014adam} is used as the optimization  method. Other hyper-parameters used are: learning rate ($0.01$), dropout probability \cite{hinton2012improving}: ($0.5$), CNN feature width ($2$), batch size ($50$), epochs ($30$) and size of the hidden state vectors ($300$). This optimal hyper-parameter values are same for the attention based RCNN and GRU encoder. We train two different question encoders trained on \textit{Semantic SQuAD} and \textit{Quora} datasets. We could not train the question encoder on the \textbf{POQR dataset} because of the unavailability of sufficient amount of similar question pairs in this dataset. Instead we use the question encoder trained on the Quora dataset to encode the questions from POQR dataset.\\

\noindent \textbf{Question Classification Network:}
To train the model we manually label (using $3$ English proficient annotators with an inter-annotator agreement of $87.98\%$) a total of $5,162$ questions\footnote{$4,000$ questions are a part of the training set of Semantic SQuAD. Remaining $1,162$ questions are from the dataset used in \newcite{liroth2002}} with their coarse and fine classes, as proposed in Section \ref{sec:taxonomy}. We release this question classification dataset to the research community. We evaluate the performance of question classification for 5-fold cross-validation in terms of F-Score. Our evaluation shows that we achieve $94.72\%$ and $86.19\%$ F-Score on coarse class ($6$-labels) and fine class ($72$-labels), respectively. We use this trained model to obtain the coarse and fine classes of questions in all datasets.\\
\indent We perform the SQC experiments with SVM classifier. We use \textit{libsvm} implementation \cite{chang2011libsvm} with linear kernel and polynomial kernel of degree $\in \{2,3,4\}$. Best performance was obtained using linear kernel. Due to the nature of POQR dataset as described in Section $4.1.2$ in the paper we employ SVM$^{light}$ \footnote{http://svmlight.joachims.org/} implementation of ranking SVMs, with a linear kernel keeping standard parameters intact. In our experiments, we use the pre-trained Google embeddings provided by \cite{mikolov2013efficient}. The focus embedding is obtained through word vector composition (averaging).

\section{Results and Analysis}\label{sec:results}
\subsection{ Results}\label{subsec:results}
We present the detailed results of semantic question ranking experiment on the Semantic SQuAD dataset in Table \ref{tab:test_results-SQUAD}. In Table \ref{tab:result:poqr}, Table \ref{tab:test_results-SQUAD} and Table \ref{tab:result-classification} we report the performnce results on the respective dataset using the models \textbf{GRU}, \textbf{RCNN, GRU-Attention and RCNN-Attention} (c.f. Section \ref{subsec:QEModel}). For all these models the results reported in the tables are based on the cosine similarity of the respective question encoder. While we introduce attention, evaluation shows improved performance for 
question encoder. 
The attention based model obatins the maximum gains of $2.40\%$ and $2.60\%$ in terms of recall and MRR for the \textit{GRU} model. The taxonomy augmented model outperforms the respective baselines and state-of-the-art deep learning based question encoder models. We obtain the best improvements for the \textit{Tax+RCNN-Attention} model, $3.75\%$ and $4.15\%$ in terms of Recall and MRR, respectively. Experiments show that taxonomy features assist in consistently improving the R@k and MRR/MAP across all the models.\\

\begin{table}[!h]
\centering
\resizebox{0.75\linewidth}{!}{%
\begin{tabular}{c|ccc|ccc}
\hline
\multirow{2}{*}{\textbf{Models}} & \multicolumn{3}{c|}{\textbf{Simple}} & \multicolumn{3}{c}{\textbf{Complex}} \\ \cline{2-7} 
 & \textbf{Simple-1} & \textbf{Simple-2}&\textbf{Overall} & \textbf{Complex-1} & \textbf{Complex-2} &\textbf{Overall}\\ \hline
GRU \cite{lei-EtAl:2016:N16-1} & 74.20 & 73.68& 73.94& 74.67 & 75.22&74.94 \\
RCNN \cite{lei-EtAl:2016:N16-1} & 76.19 & 75.81&76.00 & 75.33 & 76.44 &75.88\\ 
GRU-Attention & 75.39 & 74.83 &75.11 & 76.22 & 76.18&76.20 \\ 
RCNN-Attention & 77.28 & 77.01 &77.14 & 76.63 & 77.31&76.97 \\ \midrule
\multicolumn{7}{c}{\textbf{DNN + Taxonomy based Features}} \\ 
\midrule
Tax+GRU & 78.29 & 79.01 &78.65 & 77.63 & 78.97&78.30 \\ 
Tax+RCNN & 80.92 & 81.55 & 81.23& 80.15 & 80.83&80.49 \\
Tax+GRU-Attention & 81.69 & 81.03 &81.36  & 81.22 & 81.56 & 81.39 \\ 
Tax+RCNN-Attention & 83.67 & 83.98 & 83.82&83.32 & 84.10&83.71 \\ 
\midrule
\multicolumn{7}{c}{\textbf{State-of-the art techniques}} \\ 
\midrule
\begin{tabular}[c]{@{}c@{}} Unsupervised \textit{Cos} \\ \cite{bunescu2010learning}\end{tabular} & - & - & $73.70$& - & -&$72.60$\\ 
\begin{tabular}[c]{@{}c@{}}Supervised \textit{SVM} \\ \cite{bunescu2010learning}\end{tabular} & - & - & $82.10$&- & -& $82.50$\\ \hline
\end{tabular}
}
\caption{Semantic question ranking performance of various models on \textbf{POQR datasets}. All the numbers reported are in terms of accuracy.}
\label{tab:result:poqr}
\end{table}

\indent Performance of the proposed model on POQR dataset are shown in Table \ref{tab:result:poqr}. The `\textit{overall}' column in Table \ref{tab:result:poqr} shows the performance average on simple-1,2 and complex-1,2 datasets. We obtain improvements (maximum of $1.55\%$ with \textit{GRU-Attention} model on Complex-1 dataset) in each model by introducing attention mechanism on both simple and complex datasets. The augmentation of taxonomy features helps in improving the performance further ($8.75\%$ with \textit{Tax+RCNN-Attention} model on Simple dataset). \\
\indent The system performance on semantic question classification (SQC) experiment with Semantic SQuAD and Quora datasets are shown in Table \ref{tab:result-classification}. Similar to ranking results, we obtain significant improvement by introducing attention mechanism and augmenting the taxonomy features on both the datasets. 

\begin{minipage}{\textwidth}
  \begin{minipage}[b]{0.45\textwidth}
\resizebox{\linewidth}{!}{
\begin{tabular}{*{5}{c}}
\toprule
\textbf{\textbf{Models}} & \textbf{\textbf{R@1}} & \textbf{\textbf{R@3}} & \textbf{\textbf{R@5}} & \textbf{\textbf{MRR/MAP}} \\ 
\midrule
\multicolumn{5}{c}{\textbf{IR based Baselines}} \\ 
\midrule
TF-IDF &  54.75	&66.15&	70.25 & 61.28\\ 
Jaccard Similarity &  48.95 &	62.80 &	67.40 &57.26 \\ 
BM-25 & 56.40&	69.35	&71.45 & 61.93\\ 
\midrule
\multicolumn{5}{c}{\textbf{Deep Neural Network (DNN) based Techniques}} \\ 
\midrule
GRU \cite{lei-EtAl:2016:N16-1} & 73.25 & 84.12 & 86.39 & 76.77\\ 
RCNN \cite{lei-EtAl:2016:N16-1} & 75.10 & 86.35 & 89.01 & 78.24\\ 
GRU-Attention & 74.89 & 86.02 & 88.47 & 78.30\\ 
RCNN-Attention & 76.41 & 88.41 & 91.78 & 80.28\\ 
\midrule
\multicolumn{5}{c}{\textbf{DNN + Taxonomy based Features}} \\ 
\midrule
Tax + GRU &  76.19 & 87.02 & 88.47 & 78.98\\ 
Tax +RCNN & 78.32 & 88.91 & 92.35 & 81.49\\
Tax + GRU-Attention  & 77.35 & 89.22 & 91.28 & 80.95\\ 
Tax + RCNN-Attention & 78.88 & 90.20 & 93.25 & 83.12\\
\bottomrule
\end{tabular}
}
\captionof{table}{\textbf{Semantic Question Ranking (SQR)} performance of various models on \textbf{Semantic SQuAD} dataset, R@k and Tax denote the recall@k \& augmentation of taxonomy features.\\}
\label{tab:test_results-SQUAD}
     \end{minipage}
  \hfill
  \begin{minipage}[b]{0.50\textwidth}
    \centering
   \resizebox{\linewidth}{!}{%
\begin{tabular}{ccc}
\toprule
\textbf{Models} & \textbf{Semantic SQuAD Dataset} & \textbf{Quora Dataset}\\ 
\midrule
\multicolumn{3}{c}{\textbf{IR based Baselines}} \\
\midrule
TF-IDF &  59.28& 70.19  \\ 
Jaccard Similarity &  55.76 & 67.11  \\ 
BM-25 & 63.78 & 73.27 \\ 
\midrule
\multicolumn{3}{c}{\textbf{Deep Neural network (DNN) based Techniques}} \\ 
\midrule
GRU \cite{lei-EtAl:2016:N16-1}&74.05  &77.53 \\ 
RCNN \cite{lei-EtAl:2016:N16-1} & 77.54  & 79.32   \\
GRU-Attention & 75.18 & 79.22  \\ 
RCNN-Attention & 79.94 & 80.79\\ 
\midrule
\multicolumn{3}{c}{\textbf{ DNN + Taxonomy based Features }} \\
\midrule
Tax + GRU & 77.32 & 79.21 \\ 
Tax + RCNN & 79.89 & 81.15  \\
Tax + GRU-Attention & 78.11  & 80.91  \\
Tax + RCNN-Attention &  82.25 & 83.17 \\ 
\bottomrule
\end{tabular}
}
\captionof{table}{\textbf{Performance of Semantic Question Classification (SQC)} for various models on \textbf{Semantic SQuAD} and \textbf{Quora} datasets.\\ \\}\label{tab:result-classification}
\end{minipage}
\end{minipage}
\begin{table}[H]
\centering
\resizebox{0.8\linewidth}{!}{
\begin{tabular}{lccccc}
\hline
\textbf{Sr. No.} & \textbf{Datasets} & \textbf{All} & \textbf{-CC} & \textbf{-FC} & \textbf{-Focus Word} \\ \hline
1 & \begin{tabular}[c]{@{}c@{}}Semantic SQuAD (SQR)\end{tabular} & 83.12 & 81.66 & 81.84 &  82.20 \\ 
2 & \begin{tabular}[c]{@{}c@{}}Semantic SQuAD (SQC)\end{tabular} & 82.25 & 80.85 & 81.19  & 81.13 \\ 
3 & POQR-Simple &83.82  & 80.85  & 81.44 & 82.57 \\ 
4 & POQR-Complex & 83.71 &81.04  & 81.97  & 82.19  \\
5 & Quora & 83.17 & 80.93 & 81.75  &82.24  \\ \hline
\end{tabular}
}
\label{tab:feature_abl}
\caption{Results of feature ablation on all datasets. \textbf{SQR} results are in \textbf{MAP}. The others results are shown in terms of \textbf{Accuracy}.}
\end{table}
\subsection{Qualitative Analysis}
We analyze the results that we obtain by studying the following effects: \\
\textbf{(1) Effect of Attention Mechanism:} We analyze hidden state representation the model is attending to when deciding the semantic similarity. We depict the visualization (in appendix) of attention weight between two semantically similar question from Semantic SQuAD dataset. We observe that the improvement due to the attention mechanism in Quora dataset is comparatively less than the Semantic SQuAD dataset. The question pairs from Quora dataset have matching words, and the problem is more focused on difference rather than similar or related words.  For example, for the questions ``\textit{How magnets are made?}" and ``\textit{What are magnets made of?}", the key difference is question words `how' versus `what', while the remaining words are similar.\\
\textbf{(2) Effect of Taxonomy Features:} We perform feature ablation study on all the datasets to analyze the impact of each taxonomy feature. Table 4 shows the results\footnote{The results are statistically significant as $p <0.002$.} with the full features and after removing coarse class (-CC), fine class (-FC) and the focus features one by one. We observe from Quora dataset that the starting word of the questions \textit{(what, why, how etc.)} is a deciding factor for semantic similarity. As the taxonomy features categorize these questions into different coarse and fine classes, therefore, it helps the system in distinguishing between semantically similar and dissimilar questions. It can be observed from the results that the augmentation of CC and FC features significantly improves the performance especially on Quora dataset. Similar trends were also observed on the other datasets.
\subsection{Comparison to State-of-the-Art}
We compare the system performance on POQR dataset with state-of-the-art work of \newcite{bunescu2010learning}. \newcite{bunescu2010learning} used several cosine similarities as features obtained using bag-of-words, dependency tree, focus, main verb etc. Compared to \newcite{bunescu2010learning}, our model achieves better performance with an improvement of $2.1\%$ and $1.46\%$ on simple and complex dataset, respectively. A direct comparison to SemEval-2017 Task-3\footnote{http://alt.qcri.org/semeval2017/task3/} CQA or AskUbuntu \cite{lei-EtAl:2016:N16-1} datasets could not be made due to the difference in the nature of questions. The proposed classification method is designed for the well-formed English questions and could not be applied to multi-sentence / ill-formed questions. We evaluate \cite{lei-EtAl:2016:N16-1}'s model (RCNN) on our datasets, and report the results in Section \ref{subsec:results}. Quora has not yet released any official test set. Hence, we report the performance of 3-fold cross validation on the entire dataset to minimize the variance. We can not directly make any comparisons with others due to the non-availability of an official gold standard test set.
\subsection{Error Analysis}\label{subsec:error}
We observe the following as major sources of errors in the proposed system: \textbf{(1)} Misclassification at the fine class level is often propagated to semantic question classifier when some of the questions contain more than one sentence. For e.g., ``\textit{What's the history behind human names? Do non-human species use names?}''.
\textbf{(2)} Semantically dissimilar questions having same function words but different coarse and fine class were incorrectly predicted as similar questions. It is because of the high similarity in the question vector and focus, which forces the classifier to commit mistakes. \textbf{(3)} In semantic question ranking (SQR) task, some of the questions with higher lexical similarity to the reference question are selected in prior to the actual similar question.
\section{Conclusion}
In this work, we have proposed an efficient model for semantic question matching where deep learning models are combined with pivotal features obtained from a taxonomy. We have created a two-layered taxonomy (coarse and fine) to organize the questions in interest, and proposed a deep learning based question classifier to classify the questions. We have established the usefulness of our taxonomy on two different tasks (SQR and SQC) on four different datasets. We have empirically established that effective usage of semantic classification and focus of questions help in improving the performance of various benchmark datasets for semantic question matching.  

Future work includes building more efficient question encoders, and handling community forum questions, which are often ill-formed, using taxonomy based features.
\section{Acknowledgements}
\indent We acknowledge the partial support of Accenture IIT AI Lab. We also thank the reviewers for their insightful comments. Asif Ekbal acknowledges Young Faculty Research Fellowship (YFRF), supported by Visvesvaraya PhD scheme for Electronics and IT, Ministry of Electronics and Information Technology (MeitY), Government of India, being implemented by Digital India Corporation (formerly Media Lab Asia).

\pagebreak
\begin{appendices}
\section{Proposed Taxonomy Table}
\begin{table}[h]
\centering
\resizebox{.96\linewidth}{!}{%
\begin{tabular}{|l|l|l|l|}
\hline
\textbf{} & \textbf{Coarse Classes} & \textbf{Fine Classes} & \multicolumn{1}{c|}{\textbf{Example}} \\ \hline
\multirow{45}{*}{\textbf{Non-Decision}} & \multirow{11}{*}{\textbf{Quantification}} & Temperature & What are the approximate temperatures that can be delivered by phase change materials? \\ \cline{3-4} 
 &  & Time/Duration & How long did Baena worked for the Schwarzenegger/Shriver family? \\ \cline{3-4} 
 &  & Mass & What is the weight in pounds of each of Schwarzenegger 's Hummers? \\ \cline{3-4} 
 &  & Number & How many students are in New York City public schools? \\ \cline{3-4} 
 &  & Distance & How many miles away from London is Plymouth? \\ \cline{3-4} 
 &  & Money & What is the cost to build Cornell Tech? \\ \cline{3-4} 
 &  & Speed & Give the average speed of 1987 solar powered car winner? \\ \cline{3-4} 
 &  & Size & How large is Notre Dame in acres? \\ \cline{3-4} 
 &  & Percent & What is the college graduation percentage among Manhattan residents? \\ \cline{3-4} 
 &  & Age & How old was Schwarzenegger when he won Mr. Universe? \\ \cline{3-4} 
 &  & Rank/Rating & What rank did iPod achieve among various computer products in 2006? \\ \cline{2-4} 
 & \multirow{25}{*}{\textbf{Entity}} & Person & Who served as Plymouth 's mayor in 1993? \\ \cline{3-4} 
 &  & Location & In what city does Plymouth 's ferry to Spain terminate? \\ \cline{3-4} 
 &  & Organization & Who did Apple partner with to monitor its labor policies? \\ \cline{3-4} 
 &  & Animal & Which animal serves as a symbol throughout the book? \\ \cline{3-4} 
 &  & Flora & What is one aquatic plant that remains submerged? \\ \cline{3-4} 
 &  & Entertainment & What album caused a lawsuit to be filed in 2001? \\ \cline{3-4} 
 &  & Food & What type of food is NYC 's leading food export? \\ \cline{3-4} 
 &  & Abbreviation & What does AI stand for? \\ \cline{3-4} 
 &  & Technique & What is an example of a passive solar technique? \\ \cline{3-4} 
 &  & Language & What language is used in Macedonia? \\ \cline{3-4} 
 &  & Monuments & Which art museum does Notre Dame administer? \\ \cline{3-4} 
 &  & Activity/Process & What was the name of another activity like the Crusades occurring on the Iberian peninsula? \\ \cline{3-4} 
 &  & Disease & What kind of pain did Phillips endure? \\ \cline{3-4} 
 &  & Award/Title & Which prize did Frederick Buechner create? \\ \cline{3-4} 
  &  & Date & When was the telephone invented?\\ \cline{3-4} 
 &  & Event & What event in the novel was heavily criticized for being a plot device? \\ \cline{3-4} 
 &  & Sport/Game & Twilight Princess uses the control setup first employed in which previous game? \\ \cline{3-4} 
 &  & Policy & What movement in the '60s did the novel help spark? \\ \cline{3-4} 
 &  & Publication & Which book was credited with sparking the US Civil War? \\ \cline{3-4} 
 &  & Body & What was the Executive Council an alternate name for? \\ \cline{3-4} 
 &  & Thing & What is the name of the aircraft circling the globe in 2015 via solar power? \\ \cline{3-4} 
 &  & Feature/Attribute & What part of the iPod is needed to communicate with peripherals? \\ \cline{3-4} 
 &  & Industry Sector & In which industry did the iPod have a major impact? \\ \cline{3-4} 
 &  & Website & Which website criticized Apple 's battery life claims? \\ \cline{3-4} 
 &  & Other Tangible & In what body of water do the rivers Tamar and Plym converge? \\ \cline{3-4} 
 &  & Other Intangible & The French words Notre Dame du Lac translate to what in English? \\ \cline{2-4} 
 & \multirow{2}{*}{\textbf{Definition}} & Person & Who was Abraham Lincoln? \\ \cline{3-4} 
 &  & Entity & What is a solar cell? \\ \cline{2-4} 
 & \multirow{6}{*}{\textbf{Description}} & Reason & Why are salts good for thermal storage? \\ \cline{3-4} 
 &  & Mechanism & How do the BBC 's non-domestic channels generate revenue? \\ \cline{3-4} 
 &  & Cause \& Effect & What caused Notre Dame to become notable in the early 20th century? \\ \cline{3-4} 
 &  & Compare \& Contrast & What was not developing as fast as other Soviet Republics? \\ \cline{3-4} 
 &  & Describe & What do greenhouses do with solar energy? \\ \cline{3-4} 
 &  & Analysis & How did the critics view the movie , '' The Fighting Temptations ''? \\ \cline{2-4} 
 & \textbf{List} & \begin{tabular}[c]{@{}l@{}}Set of fine classes listed in the coarse\\ classes \textit{Quantification} and \textit{Entity}\end{tabular} & \begin{tabular}[c]{@{}l@{}} What are some examples of phase change materials? \\ Which two national basketball teams play in NYC? \end{tabular} \\ \hline
\multirow{2}{*}{\textbf{Decision}} & \multirow{2}{*}{\textbf{Selection}} & Alternative/Choice & Are the Ewell 's considered rich or poor? \\ \cline{3-4} 
 &  & True/False & Is the Apple SDK available to third-party game publishers? \\ \hline
\end{tabular}%
}
\caption{The exemplar description of proposed taxonomy classes}
\label{tax:example}
\end{table}

\section{Algorithms}

\begin{algorithm}
\begin{algorithmic}
\Procedure{Question Word}{QuesTokens}
\State \textit{WhTags} $\gets$ [WDT, WP, WP\$, WRB]
\State \textit{VbTags} $\gets$ [VB, VBD, VBP, VBZ]
\For {\textit{t} $\in$ QuesTokens}
  \If {\textit{t}.POS $\in$ \textit{WhTags}}
      \Return \textit{t}
  \EndIf
\EndFor
\For {\textit{t} $\in$ QuesTokens}
  \If {\textit{t}.POS $\in$ \textit{VbTags}}
      \Return \textit{t}
  \EndIf
\EndFor
\EndProcedure
\end{algorithmic}
\caption{Question word extraction}
\label{algo:ques_word}
\end{algorithm}

\begin{algorithm}[!thb]
\begin{algorithmic}
\Procedure{Focus}{QuesTokens}
\State \textit{qw} $\gets$ QUESTION WORD (QuesTokens)
\State \textit{depP} $\gets$ DependencyParse(QuesTokens)

\If {\textit{qw} is `how'}
	\State \Return tail of `advmod' of \textit{qw}
\EndIf
    
\If {\textit{qw}.POS is V*}
	\State \textit{obj} $\gets$ OBJECT(QuesTokens, \textit{qw})
	\State \Return \textit{obj}	
\EndIf

\If {\textit{qw}.POS is WH*}
	\If {`root' is \textit{qw}}
		\State \textit{nsubj} $\gets$ tail of `nsubj' of \textit{qw}        
		\State \Return \textit{nsubj}
	\Else
		\State \textit{obj} $\gets$ OBJECT (QuesTokens, `root')
		\State \Return \textit{obj}
	\EndIf
\EndIf

\State \Return $\textless$unk$\textgreater$
\EndProcedure
\end{algorithmic}
\caption{Focus Word Extraction}
\label{algo:focus}
\end{algorithm}

\begin{algorithm}[!htb]
\begin{algorithmic}
\Procedure{Object}{QuesTokens, \textit{qw}}

\State \textit{depP} $\gets$ DependencyParse(QuesTokens)

\State \textit{obj} $\gets$ tail of `det' of \textit{qw}
\If {\textit{obj} not NULL} \Return \textit{obj}
\EndIf    
   
\State \textit{obj} $\gets$ tail of `dobj' of \textit{qw}
\If {\textit{obj} not NULL} \Return \textit{obj}
\EndIf

\State \textit{qw} $\gets$ tail of `conj:*' of \textit{qw}
\State \textit{obj} $\gets$ tail of `dobj' of \textit{qw}
\If {\textit{obj} is NULL}
	\State \textit{comp} $\gets$ tail of `ccomp'/`xcomp' of \textit{qw}
    \State \textit{obj} $\gets$ tail of `dobj' of \textit{comp}
\EndIf
\State \Return \textit{obj}

\EndProcedure
\end{algorithmic}
\caption{Object Extraction}
\label{algo:obj}
\end{algorithm}

\newpage

\section{Additional Results}

\subsection{ K-means Clustering} 
\indent The k-means clustering was performed on the question representation obtained from the best question (RCNN-Attention) encoder of $2,000$ semantic question pairs. The clustering experiment was evaluated on the test set of Semantic SQuAD dataset ($4000$ questions). The performance was evaluated using the following metric:
\begin{equation} \label{eq:clu-rec}
\texttt{Recall} = \frac{100 \times \textit{no. of SQ pairs in same cluster}}{\textit{total no. of SQ pairs}}
\end{equation}

K-means Clustering results are as follows: R@1:50.12,  R@3:62.44 and R@5:66.58.
As the number of clusters decreases \texttt{Recall} is expected to increase as there is higher likelihood of matching questions falling in the same cluster. \texttt{Recall} with $2,000$ clusters for $2,000$ SQ pairs i.e. 4,000 questions is comparable to Recall@1 as we have 2 questions per cluster on average, \texttt{Recall} with $1,000$ clusters is a proxy for Recall@3 and \texttt{Recall} with 667 clusters is comparable to Recall@5. \\

\subsection{Semantic question classification (SQC) using IR-based Similarity}
We have used TF-IDF, BM-25 and Jaccard similarity to classify a pair of question to similar or non-similar. We calculate the score between the question using the said algorithms thereafter a optimal thresholds are used to label a question pair as `matching' or `non-matching'. If the similarity score is greater than or equal to the threshold value we set the label `matching' otherwise `non-matching'. The optimal threshold value are calculated using the validation data. The optimal threshold value are given in the table \ref{IR:threshold}.
\begin{table}[!h]
\centering
\begin{tabular}{|l|l|l|l|}
\hline
\diagbox{\textbf{Dataset}}{\textbf{Algorithm}}
                 & \textbf{TF-IDF} & \textbf{BM-25} & \textbf{Jaccard Similarity} \\ \hline
\textbf{Semantic SQuAD Dataset}              & 0.72            & 12.98          & 0.29                        \\ \hline
\textbf{Quora Dataset}            & 0.79            & 13.18          & 0.56                       \\  \hline
\end{tabular}
\caption{IR based Optimal threshold value for each dataset }
\label{IR:threshold}
\end{table}

\subsection{Attention Visualizations}
\begin{figure}[!h] 
\centering
\begin{subfigure}[b]{0.48\linewidth}
\centering
\includegraphics[width=\linewidth]{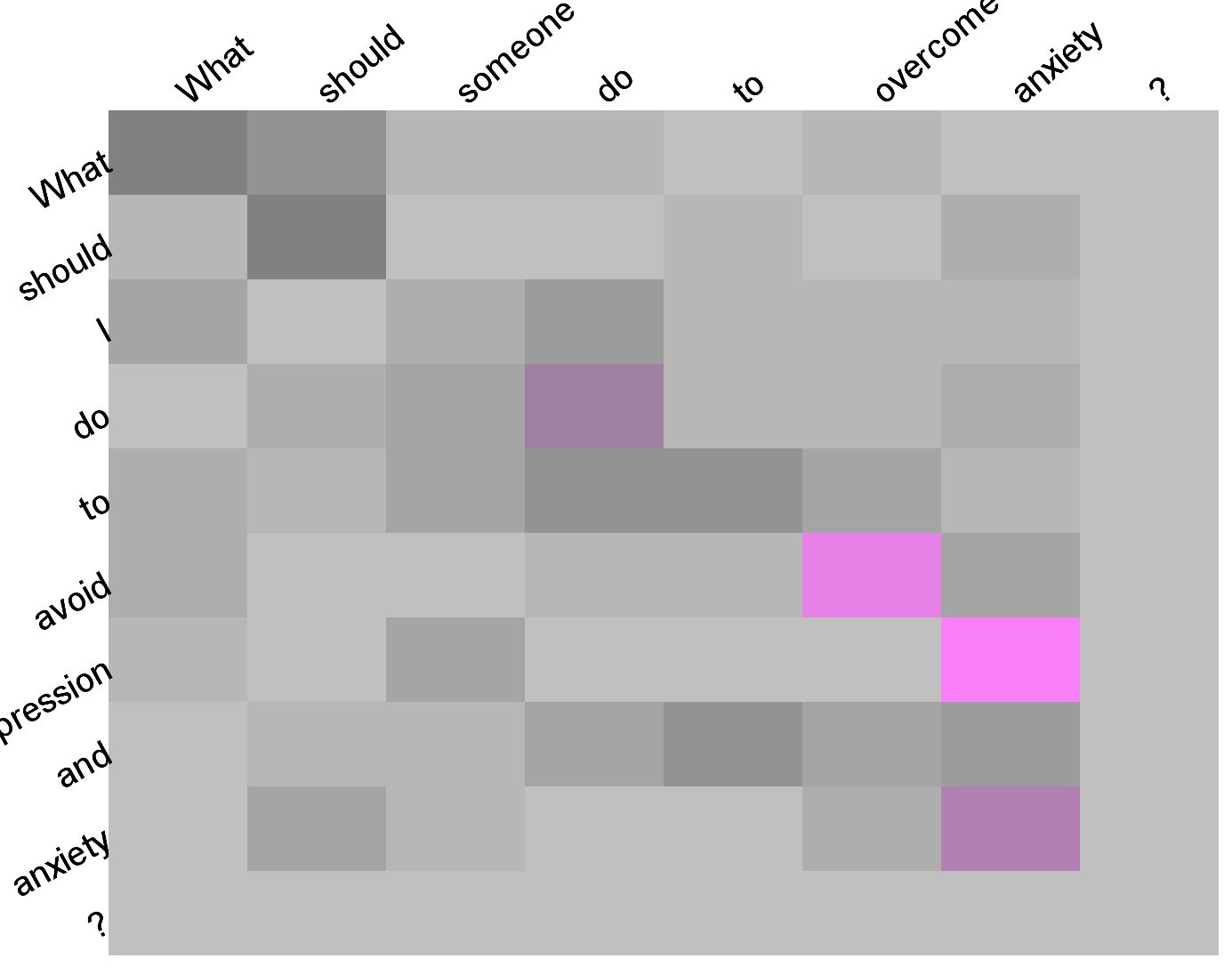}
\caption{}
\label{hindi}
\end{subfigure}
\begin{subfigure}[b]{0.48\linewidth}
\centering
\includegraphics[width=\linewidth]{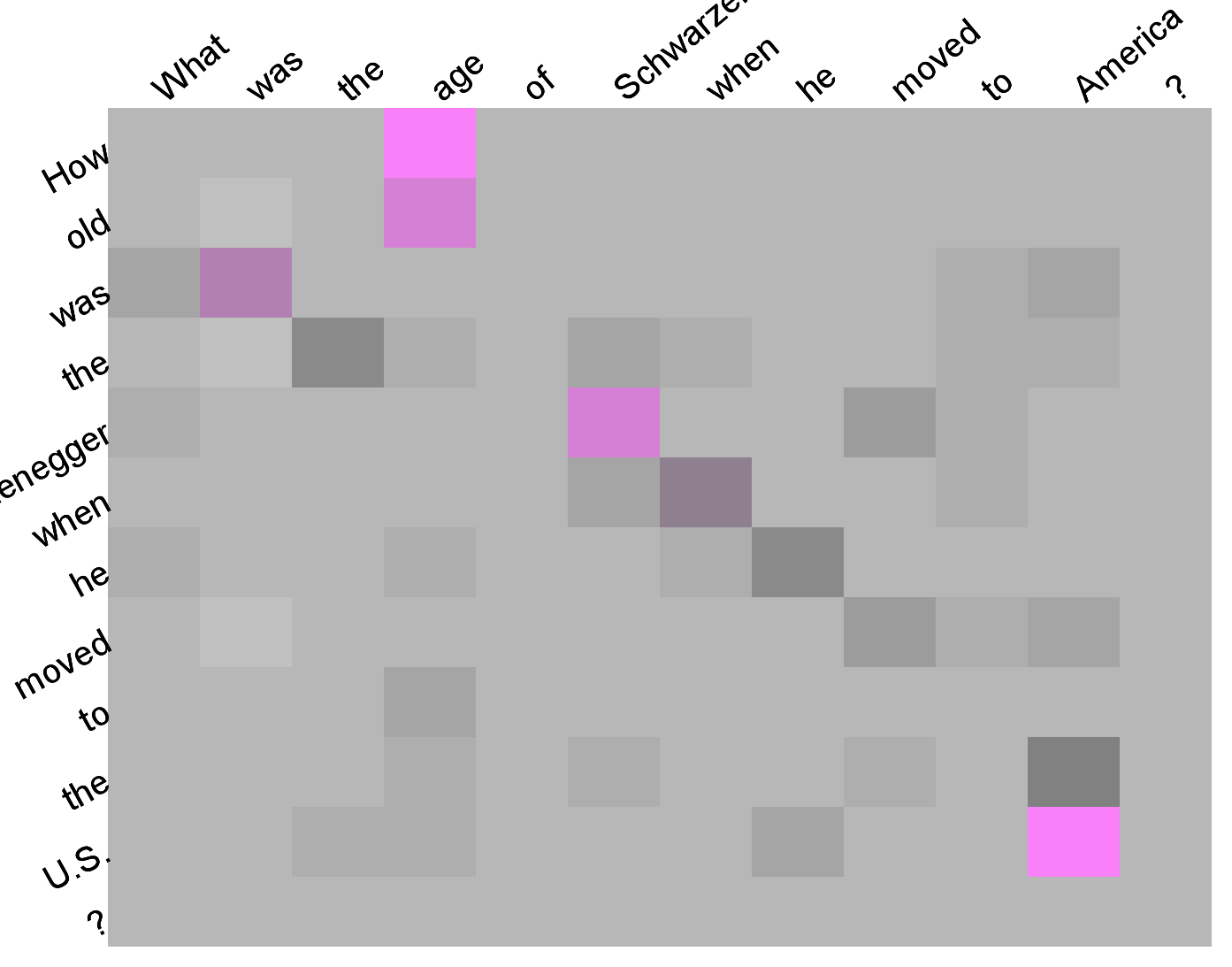}
\caption{}
\label{tamil}
\end{subfigure}
\caption{In \textbf{(a)} Attention mechanism detects semantically similar words (\textit{avoid}, \textit{overcome}). Attention mechanism is also able to align  the multi-word expression `\textit{how old}' to `\textit{age}' as shown in \textbf{(b)}}
\label{figure-1}
\end{figure}

\end{appendices}

\end{document}